\documentclass[sigconf]{acmart}

\usepackage[linesnumbered,ruled,vlined]{algorithm2e}
\usepackage{multirow}

\copyrightyear{2021}
\acmYear{2021}
\setcopyright{acmcopyright}
\acmConference[KDD '21] {Proceedings of the 27th ACM SIGKDD Conference on Knowledge Discovery and Data Mining}{August 14--18, 2021}{Virtual Event, Singapore.}
\acmBooktitle{Proceedings of the 27th ACM SIGKDD Conference on Knowledge Discovery and Data Mining (KDD '21), August 14--18, 2021, Virtual Event, Singapore}
\acmPrice{15.00}
\acmISBN{978-1-4503-8332-5/21/08}
\acmDOI{10.1145/3447548.3467358}

\begin{document}

\title{Learning Multiple Stock Trading Patterns with Temporal Routing Adaptor and Optimal Transport}
\fancyhead{}

\author{Hengxu Lin}
\authornote{The first two authors have equal contribution.}
\authornote{This work was done when the first author was an intern at Microsoft Research Asia.}
\affiliation{%
  \institution{Sun Yat-sen University}
  \city{Guangdong}
  \country{China}
}
\email{v-hengxl@microsoft.com}

\author{Dong Zhou}
\authornotemark[1]
\affiliation{%
  \institution{Microsoft Research}
  \streetaddress{No.5 Danling Street}
  \city{Beijing}
  \country{China}
  \postcode{100080}
}
\email{Zhou.Dong@microsoft.com}

\author{Weiqing Liu}
\affiliation{%
  \institution{Microsoft Research}
  \streetaddress{No.5 Danling Street}
  \city{Beijing}
  \country{China}
  \postcode{100080}
}
\email{Weiqing.Liu@microsoft.com}

\author{Jiang Bian}
\affiliation{%
  \institution{Microsoft Research}
  \streetaddress{No.5 Danling Street}
  \city{Beijing}
  \country{China}
  \postcode{100080}
}
\email{Jiang.Bian@microsoft.com}

\begin{abstract}
Successful quantitative investment usually relies on precise predictions of the future movement of the stock price. Recently, machine learning based solutions have shown their capacity to give more accurate stock prediction and become indispensable components in modern quantitative investment systems. However, the i.i.d. assumption behind existing methods is inconsistent with the existence of diverse trading patterns\footnote{In this paper, a trading \emph{pattern} means the causal relation between the available information at the current time (i.e., feature) and the stock price movement in the future (i.e., label).} in the stock market, which inevitably limits their ability to achieve better stock prediction performance.
In this paper, we propose a novel architecture, Temporal Routing Adaptor (TRA), to empower existing stock prediction models with the ability to model multiple stock trading patterns. Essentially, TRA is a lightweight module that consists of a set of independent \emph{predictors} for learning multiple patterns as well as a \emph{router} to dispatch samples to different predictors. Nevertheless, the lack of explicit pattern identifiers makes it quite challenging to train an effective TRA-based model. To tackle this challenge, we further design a learning algorithm based on Optimal Transport (OT) to obtain the optimal sample to predictor assignment and effectively optimize the router with such assignment through an auxiliary loss term.
Experiments on the real-world stock ranking task show that compared to the state-of-the-art baselines, e.g., Attention LSTM and Transformer, the proposed method can improve information coefficient (IC) from $0.053$ to $0.059$ and $0.051$ to $0.056$ respectively. Our dataset and code used in this work are publicly available\footnote{\url{https://github.com/microsoft/qlib/tree/main/examples/benchmarks/TRA}}.
\end{abstract}

\ccsdesc[500]{Computing methodologies~Machine learning}
\ccsdesc[500]{Applied computing~Forecasting}

\keywords{computational finance; stock prediction; gated network; conditional computing; optimal transport; multi-domain learning}

\maketitle

\section{Introduction}

Stock investing is one of the most popular channels for investors to pursue desirable investment goals. A successful investment usually requires precise predictions of the future movements of the stock price. Although the efficient market hypothesis (EMH)~\cite{fama1965behavior} states that predicting the stock movement is impossible given the market is efficient enough, many researchers found that the market is, in reality, less efficient than expected and can be potentially predicted. One example of well-observed predictable patterns is mean-reversion~\cite{poterba1988mean} meaning that the stock price tends to fall if it is higher than its historical average. Therefore, in some sense, the stock's future price movement can be predicted based on its historical price trend. Lately, machine learning based solutions have been employed in modern quantitative investment systems for stock prediction and prove their capacity to capture more complex patterns with powerful non-linear models from heterogeneous data sources~\cite{zhang2017stock,hu2018listening,xu2018stock,feng2018enhancing,Ding2020HierarchicalMG,Wang2020}.

To apply machine learning algorithms for stock prediction, existing approaches usually adopt a supervised learning formulation: use observable information (e.g., price, news) as features $\mathbf{x}_{i} \in \mathcal{X}$ and stock return ranks or price movement directions in the future as the label $\mathrm{y}_{i} \in \mathcal{Y}$ per stock per day respectively, then train an estimator parameterized with $\theta$ to model the underlying distribution $\mathbb{P}$. Given a collection of $N$ observations\footnote{Unless otherwise stated, in this paper we use $i$ to denote the index of stock $s$ at time $t$ for simplicity.} $\{(\mathbf{x}_{i}, \mathrm{y}_{i})\}_{i=1}^N$ as the training data, parameter $\theta$ can be effectively optimized to capture the underlying distribution by maximum likelihood estimation or empirical risk minimization. To pursue superior stock prediction performance, a variety of features extracted from diverse information sources like price~\cite{zhang2017stock,feng2018enhancing,Ding2020HierarchicalMG}, fundamental factors~\cite{chauhan2020uncertainty}, social media texts~\cite{xu2018stock,hu2018listening}, stock relation~\cite{chen2018incorporating,chen2019investment} and many model architectures like LSTM~\cite{nelson2017stock}, Attention LSTM~\cite{qin2017dual,feng2018enhancing}, Transformer~\cite{Ding2020HierarchicalMG} have been adopted and show promising gains.

Despite the success of these methods for stock prediction, they rely on the assumption that all stock samples follow an identical distribution. However, as the stock market data is a consequence of trading actions from a large number of participants, their diverse investing strategies will introduce multiple trading patterns that will be reflected in the data~\cite{ritter2003behavioral}. One such evidence is the coexistence of two contradictory phenomenons observed in stock market data: the momentum effect~\cite{fama2012size,asness2013value} (the stocks that have higher returns in the past will continue to outperform others in the future) and the reversal effect~\cite{jegadeesh1990evidence} (the stocks that have lower returns in the past may have higher returns in the future). Neglecting multiple patterns in the training samples will inevitably undermine the performance of machine learning models, which has been widely observed and discussed in other related research fields~\cite{dredze2010multi,joshi2012multi,rebuffi2017learning,rebuffi2018efficient,houlsby2019parameter}. Therefore, it is necessary to take into account the existence of multiple patterns when designing stock prediction solutions to pursue superior stock prediction performance.

In this work, we consider enhancing stock prediction by learning multiple stock trading patterns. Formally, we assume there are multiple distributions (patterns) in the stock market data $\mathbb{P} = \sum_k \nu_k \mathbb{P}_k$ with $\nu_k$ is the relative share of the $k$-th distribution, and all training and test samples come from one of these distributions\footnote{Note that if we consider collecting new test samples from an online environment, there may also be new emerging patterns. In this paper, we assume the model can be periodically updated to capture such new patterns and leave the online learning setting where the new pattern only appears during test as future work.}. In fact, if we have explicit pattern identifiers for both training and test samples, our problem can be effectively solved by Multi-Domain Learning ~\cite{dredze2010multi,rebuffi2017learning,rebuffi2018efficient,houlsby2019parameter}. However, such identifiers don't pre-exist in stock market data and are hard to determine even for human experts. Even if it is possible to distinguish different patterns for model training by using ground-truth labels, it is not applicable for the test period in which labels will not be available. Therefore, the main challenge to learn multiple trading patterns in stock market data is \emph{how to design a unified solution to discover and distinguish different patterns for both the training stage and the testing stage}.

To this end, we propose a new architecture, Temporal Routing Adaptor (TRA), as an extension module to empower existing stock prediction models with the ability to model multiple stock trading patterns. Essentially, TRA consists of a set of \emph{predictors} to model different patterns and a \emph{router} to predict which pattern a sample belongs to. The router leverages both the latent representation extracted from the backbone model (e.g., last hidden states from LSTM) as well as the temporal prediction errors of different predictors to determine a sample's pattern and assign to a specific predictor with a gating architecture~\cite{shazeer2017outrageously,hua2019,abati2020conditional,guo2020learning}. To further guarantee the discovery of diverse trading patterns, we formulate the optimal sample to predictor assignment problem as an Optimal Transport (OT) problem \cite{villani2008optimal,cuturi2013sinkhorn} and use the optimized solution from OT to guide the learning of router through an auxiliary regularization loss term.

To demonstrate the practical value of the proposed framework, we conduct extensive experiments with real-world stock market data. Experiment results show that the proposed method can bring significant gains to existing stock prediction methods. Specifically, our method can improve the information coefficient (IC) from $0.053$ to $0.059$ when compared to Attention LSTM~\cite{qin2017dual,feng2018enhancing}, and from $0.051$ to $0.056$ when compared to Transformer~\cite{Ding2020HierarchicalMG}. Further investigations demonstrate the effectiveness of different components designed in the proposed framework.

The main contributions are summarized as follows:
\begin{itemize}
    \item To the best of our knowledge, we are the first to consider designing stock prediction methods under the existence of multiple trading patterns and we also give thorough investigations of the existence and impact of multiple patterns.
    \item We propose Temporal Routing Adaptor (TRA) as an extension module to empower existing stock prediction methods with the capacity of modeling multiple stock trading patterns, and we design an effective learning algorithm based on optimal transport to further guarantee the discovery of multiple patterns.
    \item We conduct extensive experiments with real-world stock market data and compare with existing state-of-the-art baselines. Experiment results demonstrate the effectiveness of the proposed method.
\end{itemize}

\section{Related Work} \label{sec:related}
\paragraph{Stock Prediction with Deep Neural Networks}
Recently, there are many research studies to design deep learning solutions for stock prediction, which mainly fall under two categories, technical analysis (TA) and fundamental analysis (FA). TA methods focus on predicting the future movement of a stock price from market data (mainly price and volume). Among them, \cite{zhang2017stock} proposes a variant of LSTM to discover multiple frequency patterns in stock market data, \cite{qin2017dual} designs a dual-stage attention-based LSTM for stock prediction, \cite{feng2018enhancing} further improves the attention LSTM by leveraging adversarial learning, and \cite{Ding2020HierarchicalMG} also demonstrates Transformer can bring performance gains for mining long-term financial time series. FA methods are often design to capture more diversified alternative data and the most well studied data are news texts. \cite{hu2018listening} propose a Hybrid Attention Networks (HAN) to predict stock trend based on the sequence of recent related news, \cite{xu2018stock} designs a stochastic recurrent model (SRM) to address the adaptability of stock markets and \cite{cheng2020knowledge} designs a framework to leverage knowledge graph to event-based stock prediction. However, all these methods follow the identical distribution assumption and neglect the existence of multiple patterns, which limit their capacity to achieve desirable performance.

\paragraph{Multi-Domain Learning}
Multi-Domain Learning (MDL) is the field that addresses the learning challenges when the training data violates the \textit{i.i.d.} assumption and contains multiple domains (patterns)~\cite{dredze2010multi,joshi2012multi,rebuffi2017learning,rebuffi2018efficient,houlsby2019parameter}. In MDL, the \textit{domain identifiers} are assumed known during both training and test. The most straightforward approach to solve this challenge is separating the whole training set into several subsets according to their domain identifiers, and then training different models for different domains respectively. Recent studies in MDL often take the aforementioned method as the oracle baseline and mainly focus on designing more parameter-efficient network architectures, e.g., residual adaptors~\cite{rebuffi2017learning,rebuffi2018efficient,houlsby2019parameter} or depth-wise separable convolution based gating~\cite{guo2019}. However, there are no explicit domain identifiers in the stock market data and thus existing MDL solutions fall to address the challenges in this work.

\section{Multiple Trading Patterns} \label{sec:patten}
In this section, we will give several empirical evidence of the existence of multiple stock trading patterns from both investment practice and model behavior.

\begin{figure}[ht]
  \centering
   \includegraphics[width=0.8\columnwidth]{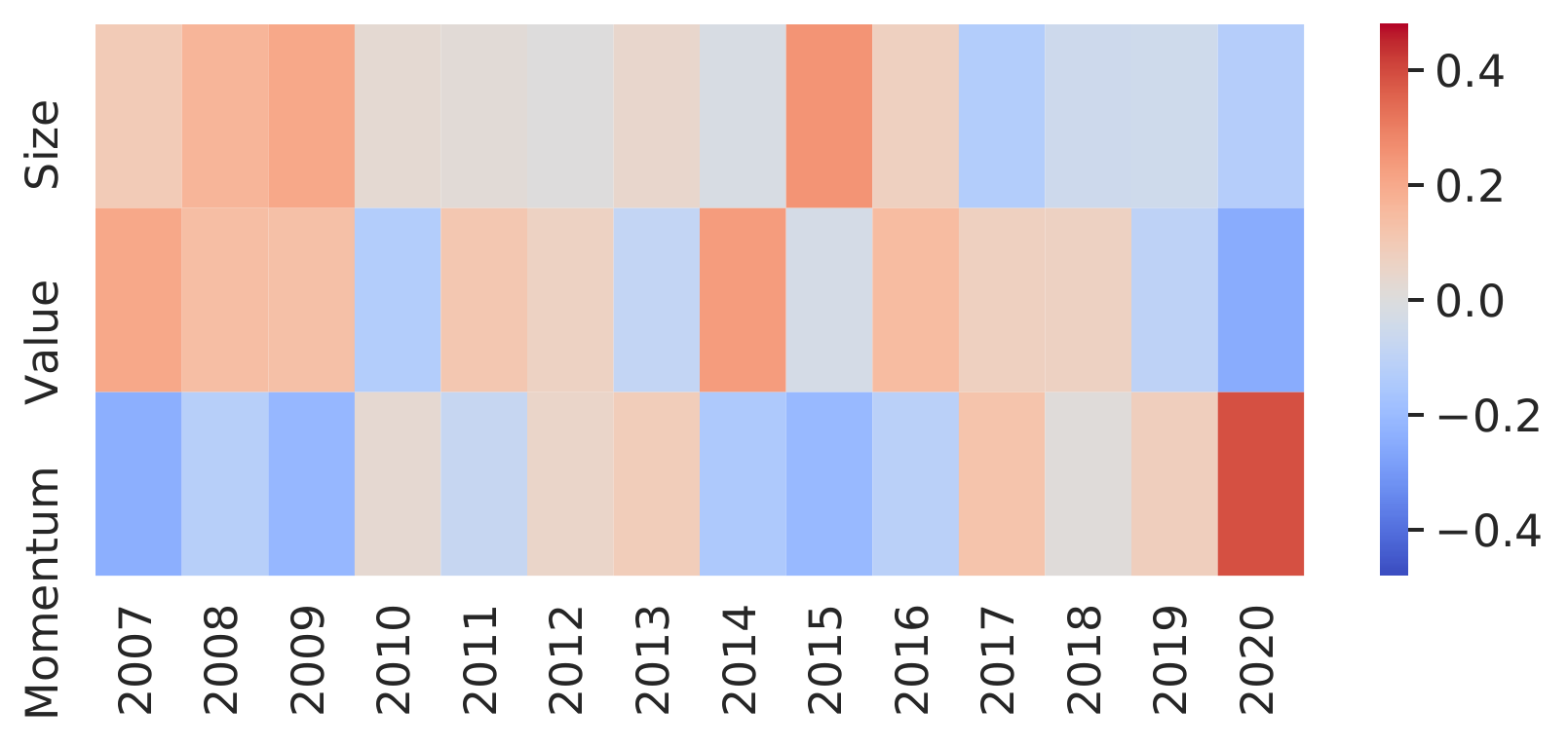}
  \caption{Annualized excess return relative to the market when equally investing in the top quintile stocks by using three investment strategies \emph{size}, \emph{value} and \emph{momentum} in China stock market.}
  \label{fig:factor}
\end{figure}

In the real world, most investors will follow specific strategies to purchase and sell stocks. When a large number of individuals adopt the same investing strategy, it will become a pattern in stock market data. Figure~\ref{fig:factor} shows the annualized excess returns relative to the market of three most known investment strategies~\cite{fama2012size}: \emph{size} (buy the small-cap stocks), \emph{value} (buy the stocks with the highest book-to-market ratio), and \emph{momentum} (buy the stocks that perform the best in the past 12 months). It can be observed that different strategies will take the lead in different periods (e.g., size before 2017, momentum since 2019), which implies there will be diverse trading patterns in the stock market data.

\begin{figure}[ht]
  \centering
   \includegraphics[width=0.8\columnwidth]{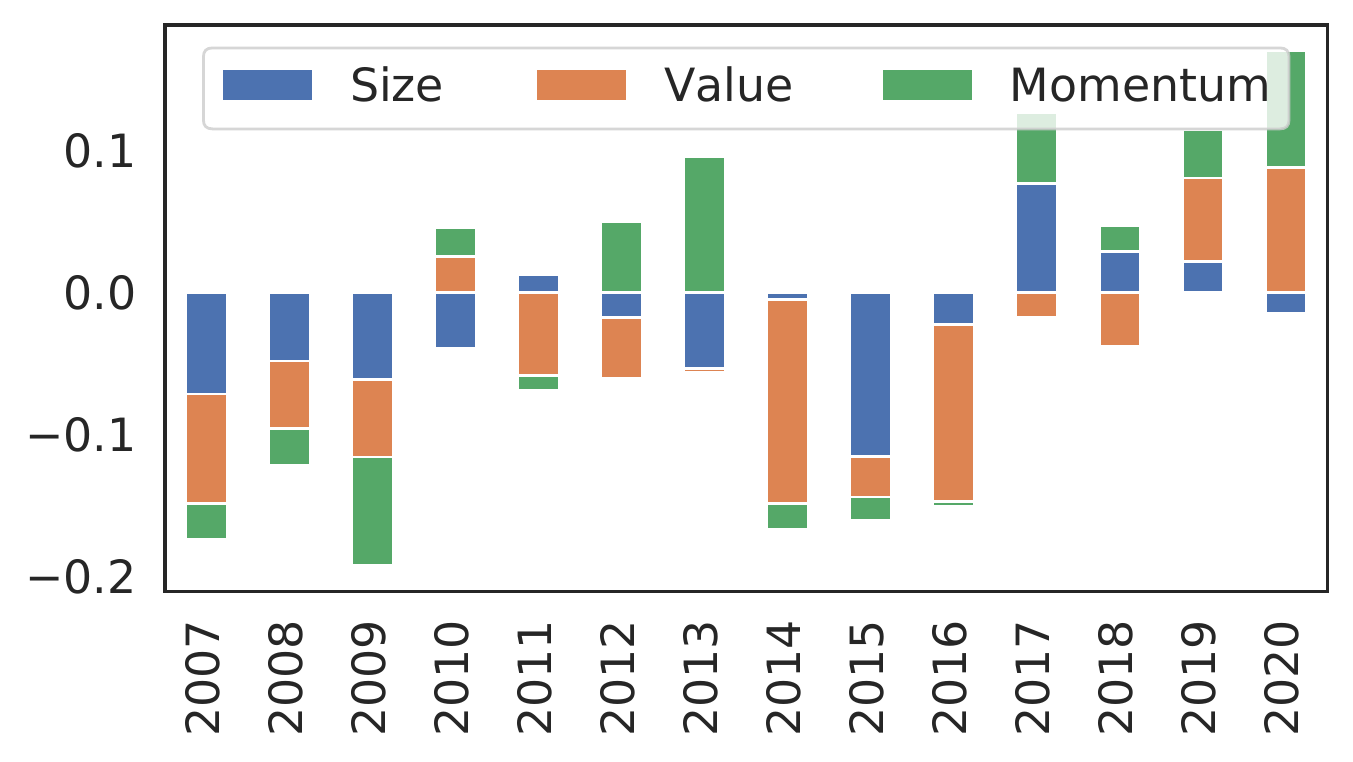}
  \caption{Linear coefficients on three features when learned with observations from different periods. Negative coefficient will be stacked below $0$. The height of the bar represents the absolute value of the coefficient.}
  \label{fig:coef}
\end{figure}

When there are multiple patterns in the data, a single model will be insufficient to capture all of them, especially when some of them are contrary to each other. In the below example, we learn multiple linear models for stock prediction with observations from different periods. We use the market capitalization rank (size), book-to-market ratio rank (value), and total return rank in the past 12 months (momentum) as features ($\mathbf{X}$), as well as the stock return ranks in the next month as the label ($\mathbf{y})$. The learned linear coefficients can be found in Figure~\ref{fig:coef} and they can be used to characterize the underlying data patterns. For example, the negative coefficient of momentum in 2009 tells us the pattern is that stock returns will be negatively correlated with their past performance. However, in 2013 the momentum coefficient becomes positive which indicates there is an opposite pattern. It will not be possible to model both these two patterns without introducing different sets of parameters to model these two patterns respectively.

\section{Problem Formulation}\label{sec:prob}
Based on the empirical observations in Section~\ref{sec:patten}, it becomes obvious that we should take into account multiple trading patterns when designing stock prediction methods. In this section, we will give a formal formulation of the learning task studied in this paper.

Let $\mathcal{X}$ denote the feature space and $\mathcal{Y}$ denote the label space. We assume data are sampled from some mixture of distributions $\mathbb{P}=\sum_k \nu_k\mathbb{P}_k$ on $\mathcal{X} \times \mathcal{Y}$, with subscript $k$ to denote the index of the $k$-th pattern and $\nu_k$ denotes its relative share. Note that both $\mathbb{P}_k$ and $\nu_k$ are unknown in the studied task. In practice, we will have a dataset containing $N$ observations $\mathcal{D}=\big\{ (\mathbf{x}_i, \mathrm{y}_i) \big\}_{i=1}^N$ sampled from the mixture of distributions $\mathbb{P}$, where an observation $\mathbf{x}_i \in \mathcal{X}$ and $\mathrm{y}_i \in \mathcal{Y}$ denote the feature and label of $i$-th sample. For model training and inference, we assume $\nu_k$ has the same proportion between the training dataset $\mathcal{D}^\mathrm{train}$ and test dataset $\mathcal{D}^\mathrm{test}$. Let $\theta$ denote the parameters that we want to learn, then the goal of learning is
\begin{equation}\label{eq:obj}
  \min_\theta \mathbb{E}_{(\mathbf{x}_i, \mathrm{y}_i) \in \mathcal{D}^\mathrm{test}} \ell(\mathbf{x}_i, \mathrm{y}_i; \theta),
\end{equation}
where $\ell$ is some measurement function.

In order to optimize $\theta$, a common practice in existing supervised learning settings is altering $\mathcal{D}^\mathrm{test}$ with $\mathcal{D}^\mathrm{train}$ in Equation~\ref{eq:obj} to optimize the parameter through empirical risk minimization. However, when there are multiple patterns in $\mathcal{D}^\mathrm{train}$, the learned single model with such an objective often performs worse than the multiple models for different patterns~\cite{rebuffi2017learning,rebuffi2018efficient,houlsby2019parameter}. Therefore, it's desired to identify different patterns and learn different sets of parameters for each pattern respectively in order to further enhance the stock prediction performance.

\section{Temporal Routing Adaptor}

In this section, we first propose the architecture of \emph{Temporal Routing Adaptor} (TRA) that can be used as an extension module to empower existing stock prediction models with the capacity to learn multiple patterns. Then we design a learning algorithm based on Optimal Transport (OT) to further guarantee the discovery of multiple patterns. Last, we will give more implementation details for effectively training TRA-based models.

\subsection{Architecture}
\begin{figure}[ht]
   \includegraphics[width=0.5\columnwidth]{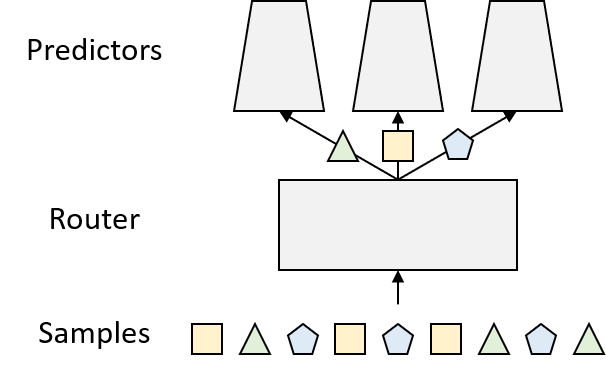}
  \caption{An overview of the proposed \emph{predictors} + \emph{router} framework. The \emph{predictors} are responsible for learning different patterns while the \emph{router} controls how samples are assigned to different predictors.}
  \label{fig:router}
\end{figure}

In order to capture multiple stock trading patterns, the model needs to has the ability to provide different parameters for samples that belongs to different patterns. In this work, we propose Temporal Routing Adaptor (TRA) with two components to fulfill this goal: a set of \emph{predictors} to learn different trading patterns, and a \emph{router} to select the appropriate predictor. As illustrated in Figure~\ref{fig:router}, the \emph{router} will assign samples with the same pattern to the same \emph{predictor}, such that during the training stage the assigned predictor will capture the corresponding pattern from the assigned samples, and during the testing stage the best matched predictor will be used for inference.

\begin{figure}
  \centering
  \includegraphics[width=1.0\columnwidth]{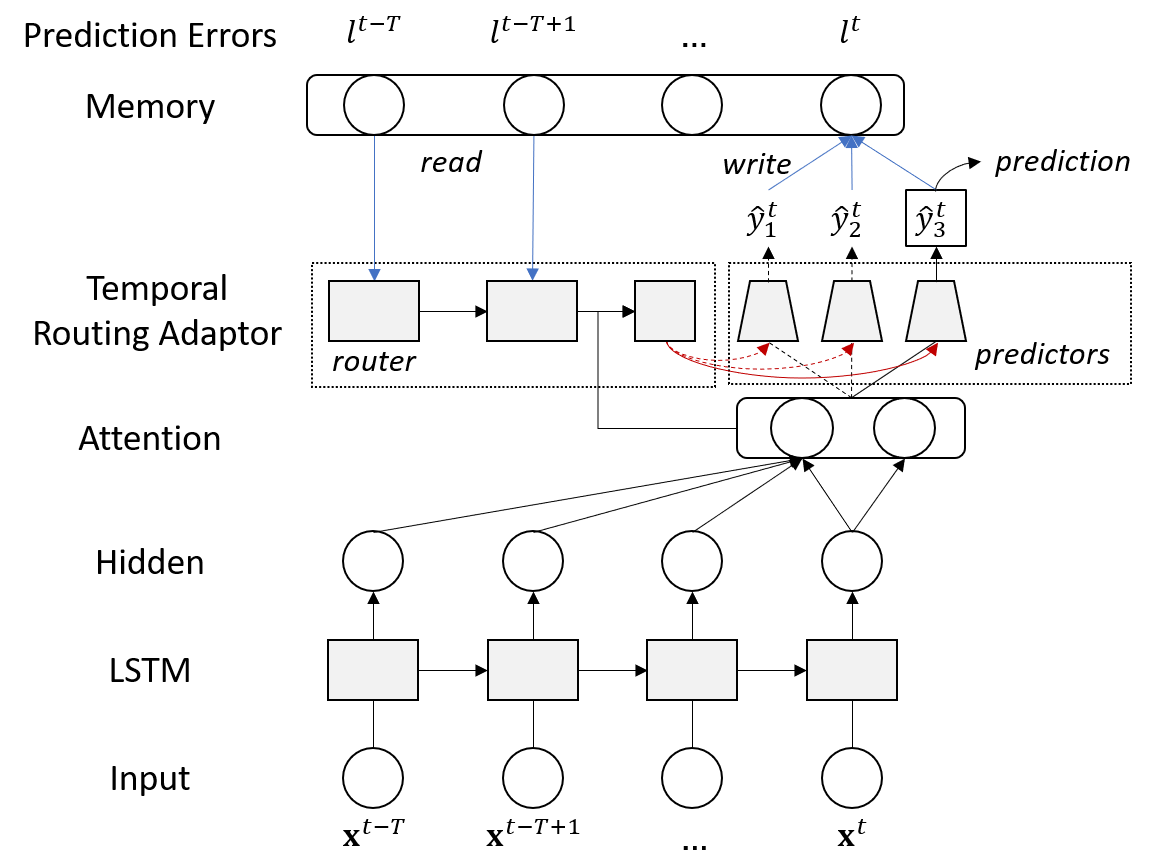}
  \caption{An example to use Temporal Routing Adaptor (TRA) as an extension module on Attention LSTM. The router uses both latent representation from Attention LSTM and temporal prediction errors stored in memory to select the best predictor for the current sample.}
  \label{fig:framework}
\end{figure}

Let's first consider the design of the \emph{predictors}. The most straightforward approach should be introducing multiple parallel models for different patterns, yet it will introduce $(K - 1)\times$ more parameters compared to the original model if we want to model $K$ different patterns. Using such many parameters will make the model more prone to over-fitting as observed in our experiments. Therefore, we only replace the output layer in the traditional stock prediction models with a set of linear predictors. This design can be easily plugged into existing stock prediction networks with negligible added parameters. We use Attention LSTM as the backbone feature extractor to demonstrate this in Figure~\ref{fig:framework}. Another benefit of using this design is that if we keep only one predictor, then TRA will become the classical Attention LSTM and thus our framework won't cause any performance regression.

For the design of the \emph{router}, it depends on what kind of information we want to use to predict the underlying pattern to which a sample belongs. In this work, we utilize two types of information to predict a sample's pattern and assign it to the appropriate predictor: 1) the latent representation from the backbone feature extractor, 2) temporal prediction errors of different predictors. The first type of information should be easy to understand: certain patterns can be directly distinguished from the latent representation $p(\mathbf{x})$. And instead of using raw features, we choose to use latent representation because it is more related to the underlying pattern $p(\mathrm{y}|\mathbf{x})$. Moreover, considering that the latent representation still cannot give enough information on $p(\mathrm{y}|\mathbf{x})$ as the patterns in stock market data are extremely hard to capture, we leverage the second type of information as a supplement.

The second type of information is inspired by the real-world investment practice: investors will periodically change their investment strategies (analogue to multiple predictors) from one to the other based on the recent performances (analogue to prediction errors) of these strategies. Merrill Lynch investment Clock\footnote{\url{https://macro-ops.com/the-investment-clock/}} is such an example that shows certain investment strategies will take the lead following the macroeconomic cycle. Therefore, we can use the temporal prediction errors of different predictors in the past days to predict which predictor should be chosen in the next day for the current sample. To this end, we use a recurrent neural network to summarize the history performance of different predictors into an embedding vector and concatenate this embedding with the latent representation as the input of the router to predict the pattern.

Formally, let $\Theta=\{ \theta_1, \theta_2, ..., \theta_K \}$ denote the parameters of $K$ predictors, $\pi$ denote the parameter of the router, and $\psi$ denote the parameter of the shared feature extractor. For the $k$-th predictor, we denote the entire model (feature extractor and predictor) to give prediction for the $i$-th sample as $\hat{\mathrm{y}}_{ik} = \theta_k \circ \psi(\mathbf{x}_i)$. We also use bold symbol $\hat{\mathbf{y}}_i = \Theta \circ \psi (\mathbf{x}_i) \triangleq [\theta_1 \circ \psi(\mathbf{x}_1), \theta_2 \circ \psi(\mathbf{x}_2), ..., \theta_K \circ \psi(\mathbf{x}_K)]^\intercal$ to denote the predictions from all predictors. The prediction errors for the $i$-th sample can be then calculated as
\begin{equation}
    l_i = \ell(\hat{\mathbf{y}_i}, \mathbf{y}),
\end{equation}
 where we use $l_i$ to denote the vectorized prediction errors $[\ell(\hat{\mathrm{y}}_1, \mathrm{y}_1)$, $\ell(\hat{\mathrm{y}}_2, \mathrm{y}_2)$, $...$, $\ell(\hat{\mathrm{y}}_K, \mathrm{y}_K)]^\intercal$ of the $i$-th sample for simplicity, and $\ell$ is some measurement function related to the stock prediction task.

Note that we can also index the $i$-th sample by the stock id $s$ and the timestamp $t$, thus we can aggregate the temporal prediction errors for this sample as
\begin{equation} \label{eq:tpe}
  \mathbf{e}_i \triangleq \mathbf{e}_{i=(s,t)} = \big[l_{(s,t-T)},l_{(s,t-T+1)},...,l_{(s,t-h)}\big]^\intercal,
\end{equation}
where $T$ is a maximized lookback window, $h$ is an extra gap to avoid using future information\footnote{$h$ should be larger than the horizon of label to avoid using future information.}. To speed up model training, we also leverage \emph{memory} to store the prediction errors. More details about the memory implementation can be found in Section~\ref{src:algo}.

Given both latent representation and temporal prediction errors, a straightforward approach for the router to give prediction is by applying a soft attention on different predictors through softmax. Denote the latent representation of sample $i$ extracted from $\psi $ as $\mathbf{h}_i=\psi(\mathbf{x}_i)$, then such softmax-based routing mechanism can be generated by
\begin{equation}\label{eq:softmax}
  \begin{aligned}
    \mathbf{a}_i &= \pi(\mathbf{h}_i, \mathbf{e}_i), \\
    \mathbf{q}_i &= \frac{\mathrm{exp}(\mathbf{a}_i)}{\mathrm{sum}(\mathrm{exp}(\mathbf{a}_i))}, \\
    \hat{\mathrm{p}}_i &= \mathbf{q}_i^\intercal \hat{\mathbf{y}}_i
  \end{aligned}
\end{equation}
where $\mathbf{q}_i$ is the routing attention and $\hat{\mathrm{p}}_i$ is the final prediction of the entire model.

However, it's important for the router to give discrete selection in order to really distinguish different patterns.  Instead of using the continuous softmax in Equation~\ref{eq:softmax}, the straight approach to achieve the discrete selection is using $argmax(\mathbf{q}_i)$, which however is not differentiable. To achieve differentiable but discrete routing, we utilize the gumbel-softmax trick~\cite{jang2016categorical} to enable the differentiability of the router. Essentially, it will add gumbel noise to the logits $\mathbf{a}_i$ and leverage the reparametrization trick to make the computation fully differentiable.
\begin{equation}\label{eq:gumbel}
  \begin{aligned}
    \mathbf{q}_i &= \frac{\mathrm{exp}((\mathbf{a}_i+ \epsilon) / \tau)}{\mathrm{sum}(\mathrm{exp}((\mathbf{a}_i+ \epsilon) / \tau))},
  \end{aligned}
\end{equation}
where $\epsilon=[\epsilon_1, \epsilon_2, ..., \epsilon_K]$ are i.i.d. sampled drawn from Gumbel(0, 1)\footnote{The Gumbel(0,1) distribution can be sampled by first drawing uniform distribution $z \sim \mathrm{Uniform}(0,1)$ then applying $-\mathrm{log}(-\mathrm{log}(z))$.} distribution and $\tau$ is the temperature that controls the sharpness of the output distribution.

\subsection{Optimal Transport}
While TRA enables the capacity of modeling multiple trading patterns in a single network, it's still challenging to discover the underlying multiple trading patterns. In fact, we observed in our experiments that directly training TRA will end up with some trivial solutions, where all samples are routed to the same predictor and thus fails to capture the diverse patterns. A straightforward approach is to directly add strong regularization to force the router give relatively balanced sample assignment to different predictors. In this work, we propose a better approach to achieve such regularization in the meanwhile encourage the discovery of multiple trading patterns.

\begin{figure}
  \centering
  \includegraphics[width=0.8\columnwidth]{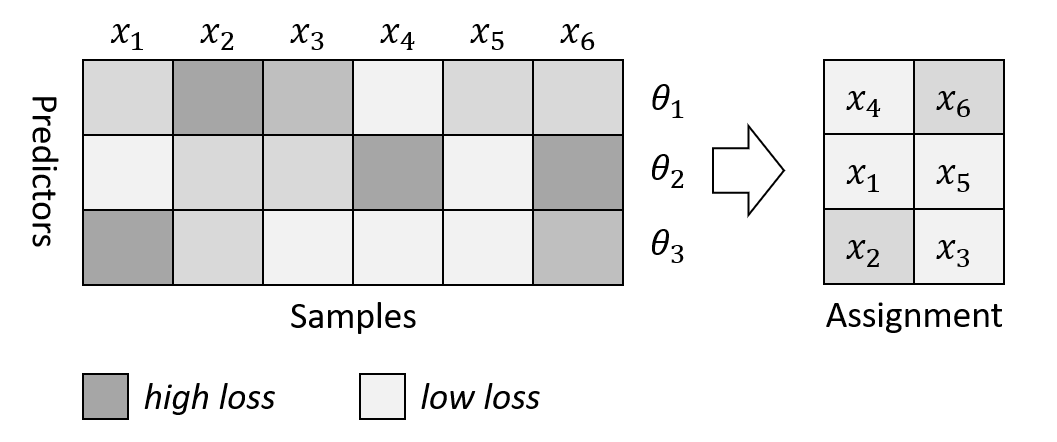}
  \caption{Our goal is assigning samples $x_1$ to $x_6$ to three predictors $\theta_1$ to $\theta_3$ with the goal of minimizing the prediction loss while keeping the number of assigned samples balanced.}
  \label{fig:transport}
\end{figure}

Our main idea is to \emph{assign a sample to the predictor that has the smallest loss (objective) while keeping the number of assigned samples to a specific predictor proportional to the relative share of the corresponding pattern (constraint)}. Figure~\ref{fig:transport} gives an illustration of our idea and we can further formalize it as an optimization problem.
First we can pack the prediction errors of all samples $\times$ all predictors into a loss matrix $\mathbf{L}$, with $\mathrm{L}_{ik}=\ell\big(\theta_k \circ \psi(\mathbf{x}_i), \mathrm{y}_i\big)$ is the prediction loss of sample $i$ when choosing predictor $\theta_k$. Denote the sample assignment matrix for $N$ samples and $K$ predictors as $\mathbf{P} \in \{0, 1\}^{N \times K}$, then we are able to get the assignment $\mathbf{P}$ by solving the following optimal transport problem~\cite{villani2008optimal,cuturi2013sinkhorn}
\begin{equation}\label{eq:transport}
  \begin{aligned}
    \min_{\mathbf{P}} & \ \langle\mathbf{P}, \mathbf{L}\rangle, \\
    s.t. & \sum_{i=1}^N \mathbf{P}_{ik} = \nu_k * N, \ \forall k=1...K \\
    & \sum_{k=1}^K \mathbf{P}_{ik} = 1, \ \forall i=1...N \\
    & \mathrm{P}_{ik} \in \{0,1\}, \ \forall i=1...N,k=1...K,
  \end{aligned}
\end{equation}
where $\langle\cdot,\cdot\rangle$ stands for the Frobenius dot-product and $\nu_k$ is the relative share of the $k$-th pattern as introduced in Section~\ref{sec:prob}. The first constraint ensures the assigned samples are proportional to the relative shares of the latent patterns, and the second to third constraints ensure the discrete choice of only one predictor. This problem is combinatorial and it will take a great amount of time to get the exact solution with the standard solvers. To overcome this challenge, we leverage the \emph{Sinkhorn's matrix scaling algorithm} ~\cite{cuturi2013sinkhorn} to obtain an approximated solution with the benefit of several orders of magnitude speed-ups.

After obtaining the optimal sample assignment matrix $\mathbf{P}$, we further investigate how to use it to assist the training of TRA. Note that the optimized sample assignment $\mathbf{P}$ depends on the prediction loss $\mathbf{L}$ and thus also depends on the unknown label, it cannot be directly used to assign samples to different predictors during the testing stage as labels will not be available. Here we consider to use it to guide the learning of the router through adding an auxiliary regularization term into the objective
\begin{equation} \label{eq:loss}
  \min_{\Theta, \pi, \psi} \mathbb{E}_{(\mathbf{x}_i, \mathrm{y}_i) \in \mathcal{D}^\mathrm{train}}\big[\ell(\mathbf{x}_i, \mathrm{y}_i; \Theta,\pi,\psi) - \lambda \textstyle \sum_{k=1}^K \mathrm{P}_{ik} \mathrm{log}(q_{ik}) \big],
\end{equation}
where $q_{ik}$ is the predicted probability from gumbel-softmax in Equation~\ref{eq:gumbel}, and $\lambda$ is a hyper-parameter controlling the regularization strength. The regularization term is defined by cross entropy with $\mathbf{P}$ as the target distribution.

Note that we don't know the exact number of trading patterns $K$, so it is treated as a hyper-parameter and will be compared by setting different values in our experiments. Besides, the relative shares $\nu_k$ for different patterns in Problem~\ref{eq:transport} are also unknown. We address this by first treating different patterns having equal shares (i.e.,  $\nu_k = \frac{1}{K}$) in the early stage of training, then we gradually decay $\lambda$ with decay rate $\rho$ during training so the router can discover more suitable pattern share in the later stage.

\subsection{Implementation}\label{src:algo}
As TRA needs temporal prediction errors to determine the assignment of a sample, the straightforward approach to implement this will require a all forward pass of all training samples in each single training step, which will make computation costs too high to bear. To this end, we introduce an external memory $\mathbf{M} \in \mathbb{R}^{N \times K}$ to cache the prediction errors for all $N$ samples with $K$ predictors calculated in previous training steps. Then we follow a two-stage procedure to update the memory: refresh the whole memory before each training epoch and refresh partial memory of mini-batch samples only appeared in the current training steps on the fly. By using this strategy we can achieve $N/\mathrm{batch size}$ speed-ups.

Algorithm~\ref{algo} lists the whole process of training TRA, with the optimal transport regularized objective and the two-stage memory updating strategy. Essentially, each training epoch consists of: 1) inference all training samples and store the prediction errors of all predictors into the memory $\mathbf{M}$; 2) sample mini batches to update the model parameters with Equation~\ref{eq:loss} as well as update the memory for samples in this mini batch. Note that it is important to strictly follow the temporal order to sample batch samples during testing, because TRA needs prediction errors from former timestamps as input information for the router to select the right predictor for the current timestamp.

\begin{algorithm}
\SetAlgoLined
	\KwIn{training data $\mathcal{D}^\mathrm{train}$, model parameters $\pi$, $\psi$, $\Theta$ (with size $K$),  hyper-parameters $\lambda$, $\rho$ }
	\KwOut{$\psi^*$, $\pi^*$, $\Theta^*$}
	Randomly initialize $\pi$, $\psi$, $\Theta$\;
	Initialize memory $\mathbf{M}$\;
	\While{not meet the stopping criteria} {
		\tcc{step 1: update the memory.}
		\For{$i \gets1 $ \KwTo $N$} {
		    Update the memory of the $i$-th sample by $\mathbf{M}_i = \ell(\Theta \circ \psi(\mathbf{x}_i), \mathbf{y}_i)$\;
		}
		\tcc{step 2: optimize model parameters.}
		\For{$\mathcal{D}^\mathrm{batch} \gets \mathcal{D}^\mathrm{train}$} {
		    Decay regularization parameter $\lambda=\lambda*\rho$\;
		    Get the loss matrix $\mathbf{L}$ for current batch $\mathcal{D}^\mathrm{batch}$\;
		    Solve the optimal transport problem in Equation~\ref{eq:transport} via Sinkhorn's algorithm\;
		    Calculate the regularized loss with Equation~\ref{eq:loss} and update $\Theta$, $\pi$, $\psi$ via gradient descent algorithms\;
		    Update memory $\mathbf{M}$ for samples in this batch\;
		}
	}
\caption{Algorithm for Training TRA.}
\label{algo}
\end{algorithm}

\section{Experiments}
To demonstrate the effectiveness of the proposed method, we further conduct extensive experiments on the real-world stock market data. In the following subsections, we will first introduce our experimental settings, including task and dataset description, baseline methods, evaluation metrics and hyper-parameter settings. Then we will give the experiment results of the compared methods. Last, we further conduct comprehensive experiments to demonstrate the effectiveness of the proposed method.

\subsection{Settings}
\subsubsection{Task.}
In this work, we study the stock ranking prediction task with labels defined as the percentiles of cross-sectional stock returns at the end of next month. Using percentiles instead of raw returns as the learning targets has the following advantages: 1) the learning process will be less affected by extraordinary return outliers; 2) percentiles are uniform distributed and not affected by the market fluctuation (which will make the data non-stationary); 3) the model trained to predict cross-section stock return percentiles can be easily used by most portfolio construction strategies like long-short portfolio. As percentiles are continuous, the stock ranking prediction will be studies as a regression task in this work.

\subsubsection{Dataset}
We construct our dataset based on China stock market data\footnote{The China stock is less mature such that the publicly available information can still be used for stock prediction. For the developed market, the public information are no longer enough to predict the market and the proprietary data sources are demanded in order to learn stock prediction models, which however are hard to obtain for the research community.}. Specifically, we collect publicly available data for CSI800 stocks\footnote{CSI800 consists of the 800 largest and most liquid China stocks. More information can be found at \url{http://www.csindex.com.cn/en/indices/index-detail/000906}.} from baostock.com. Then we extract 16 features that have been proved informative in China stock market, including market capitalization, price-to-EPS, price-to-book value, price-to-sales, price-to-net cash flow, asset turnover ratio, net profit margin, receivables turnover ratio, EPS growth, asset growth, equity growth, 12 month momentum, 1 month reversal, close to 12 month highest, close to 12 month lowest, and maximum return in last month. All features are transformed into their ranks to match the stock ranking task within each trading day.

Finally, we follow the temporal order (avoid the data leakage problem) to split data into training (from 10/30/2007 to 05/27/2016), validation (from 09/26/2016 to 05/29/2018) and test (from 09/21/2018 to 06/30/2020). Note that the extra splitting gaps (e.g., 05/29/2018 to 09/21/2018) are intentionally introduced to avoid the leakage of both feature and label, because we will use the features in last $60$ trading days for time series models and stock returns ranks in the next $21$ trading days (1 month) as the learning targets.

\subsubsection{Baselines.}
We compare our proposed method with the following baselines:
\begin{itemize}
    \item \textbf{Linear} is a linear regression model.
    \item \textbf{LightGBM}~\cite{ke2017lightgbm} is non-linear model based on gradient boosted trees.
    \item \textbf{MLP} is a non-linear model based on neural networks.
	\item \textbf{SFM}~\cite{zhang2017stock} is a redesigned recurrent neural network by decomposing the hidden states of LSTM into multiple frequency components to model multi-frequency trading patterns.
	\item \textbf{ALSTM}~\cite{qin2017dual,feng2018enhancing} adds an external attention layer into the LSTM architecture to attentively aggregate information from all hidden states in previous timestamps.
	\item \textbf{Trans.}~\cite{Ding2020HierarchicalMG} adopts the Transformer~\cite{vaswani2017attention} architecture that has been used by \cite{Ding2020HierarchicalMG} for stock prediction.
\end{itemize}

In addition to the above baselines, we also consider the following heuristic approach introduced in Section~\ref{sec:patten} to learn multiple trading patterns
\begin{itemize}
    \item \textbf{ALSTM+TS/Trans.+TS} simply assigns samples to different predictors according to which time periods they belong to. As test samples belong to a new time period and thus cannot be assigned to the learned models, we simply average predictions from all the learned models.
\end{itemize}

\subsubsection{Evaluation Metrics}
In addition to evaluating the prediction errors of stock ranking by regression metrics like Mean Square Error (MSE) and Mean Absolute Error (MAE), the dominate ranking metric used in finance is the information coefficients (IC), which can be calculated as
\begin{equation}
    \mathrm{IC}=\frac{1}{N} \frac{(\hat{\mathbf{y}}- mean(\hat{\mathbf{y}}))^\intercal (\mathbf{y} - mean(\mathbf{y}))}{std(\hat{\mathbf{y}}) * std(\mathbf{y})},
\end{equation}
where $\hat{\mathbf{y}}$ and $\mathbf{y}$ are the predicted rankings and actual rankings respectively. In practice, IC is calculated for different trading days and their average will be reported. To show the stability of IC, we also report the information ratio of IC, i.e., ICIR, which is calculated by dividing the average by the standard deviation of IC.

In addition, we also evaluate the ranking performance of the learned model by constructing a long-short portfolio by longing stocks of the top decile and shorting stocks of the bottom decile\footnote{In China stock market, shorting stocks can be achieved by securities margin trading.}. We will measure the portfolio's performance by annualized return (AR), annualized volatility (AV), annualized Sharpe Ratio (SR) and maximum drawdown (MDD). The details for portfolio construction and metric definitions can be found in in Appendix~\ref{sec:port_metrics}.

\subsubsection{Parameter Settings}
To compare the performance of different baselines on our dataset, we first search architecture parameters and hyper-parameters for \textbf{LightGBM}, \textbf{MLP}, \textbf{SFM}, \textbf{ALSTM} and \textbf{Trans.} and select the best parameter based on the performance of validation set. The architecture parameters and hyper-parameter configurations for the search space can be found in Appendix~\ref{sec:arch}. For our methods, we use the same architecture parameter as the backbone models \textbf{ALSTM} and \textbf{Trans.} for fair comparison and only search the learning rate and decay rate $\rho$ for regularization strength $\lambda$ used in Equation~\ref{eq:loss}.

\subsection{Main Results}

\begin{table*}
	\centering
	\caption{Stock ranking and portfolio performance of the compared methods. For MSE and MAE, we also report the standard deviation of 5 random seeds as $(\cdot)$. $\uparrow$ means the larger the better while $\downarrow$ means the smaller the better. }\label{tab:result}
	\begin{tabular}{c|cccc|cccc}
        \toprule
        \multirow{2}{*}{\textbf{Method}} & \multicolumn{4}{c|}{\textbf{Ranking Metrics}} & \multicolumn{4}{c}{\textbf{Portfolio Metrics}} \\
        & MSE ($\downarrow$) & MAE ($\downarrow$) & IC ($\uparrow$)& ICIR ($\uparrow$) & AR ($\uparrow$) & AV ($\downarrow$) & SR ($\uparrow$) & MDD ($\downarrow$) \\
		\hline
        Linear                    &          0.163 &          0.327 &  0.020 &  0.132 &  -3.2\% &  16.8\% &  -0.191 &  32.1\% \\
        LightGBM                      &  0.160(0.000)  &  0.323(0.000) &  0.041 &  0.292 &   7.8\% &  15.5\% &   0.503 &  25.7\% \\
        MLP                       &  0.160 (0.002) &  0.323 (0.003) &  0.037 &  0.273 &   3.7\% &  15.3\% &   0.264 &  26.2\% \\
        SFM                       &  0.159 (0.001) &  0.321 (0.001) &  0.047 &  0.381 &   7.1\% &  14.3\% &   0.497 &  22.9\% \\
        ALSTM                     &  0.158 (0.001) &  0.320 (0.001) &  0.053 &  0.419 &  12.3\% &  \textbf{13.7\%} &   0.897 &  \textbf{20.2}\% \\
        Trans.               &  0.158 (0.001) &  0.322 (0.001) &  0.051 &  0.400 &  14.5\% &  14.2\% &   1.028 &  22.5\% \\
        \hline
        ALSTM+TS                  &  0.160 (0.002) &  0.321 (0.002) &  0.039 &  0.291 &   6.7\% &  14.6\% &   0.480 &  22.3\% \\
        Trans.+TS            &  0.160 (0.004) &  0.324 (0.005) &  0.037 &  0.278 &  10.4\% &  14.7\% &   0.722 &  23.7\% \\
        \hline
        \textbf{ALSTM+TRA (Ours)}       &  \textbf{0.157 (0.000)} &  \textbf{0.318 (0.000)} &  \textbf{0.059} &  \textbf{0.460} &  12.4\% &  14.0\% &   0.885 &  20.4\% \\
        \textbf{Trans.+TRA (Ours)} &  \textbf{0.157 (0.000)} &  0.320 (0.000) &  0.056 &  0.442 &  \textbf{16.1}\% &  14.2\% &   \textbf{1.133} &  23.1\% \\
		\bottomrule
	\end{tabular}
\end{table*}

Table~\ref{tab:result} summarized the evaluation performances of all compared methods. From the exhibited results, we have the following observations:
\begin{itemize}
    \item The proposed method can be used to further improve the performance of previous state-of-the-art baselines  (\textbf{ALSTM} and \textbf{Trans.}). Specifically, compared to \textbf{ALSTM}, our method \textbf{ALSTM+TRA} can reduce the prediction errors measured by MSE from $0.158$ to $0.157$ and improve the ranking metric measured by IC from $0.053$ to $0.059$. Also compared to \textbf{Trans.}, our method \textbf{Trans.+TRA} can also improve IC from $0.051$ to $0.056$. These results demonstrate the efficacy of the proposed method to enhance stock prediction.
    \item The two heuristic methods \textbf{ALSTM+TS} and \textbf{Trans.+TS} that model multiple patterns by splitting the dataset into multiple periods even harm the model's performance. We postulate the reason is that the learned individual models can only capture the pattern of a specific period and simply averaging predictions from these models during testing will mixes up different patterns and thus performs poorly. This shows that selecting the right model for inference is also critical to ensure the performance. As a comparison, the router in the proposed TRA framework has the ability to fulfill this goal.
    \item From the portfolio metrics, it can be observed that our method can also achieve desirable performance. Compared to \textbf{Trans.}, our method achieves higher annualized return (ours $16.1$\% vs \textbf{Trans.} $14.5$\%) as well as higher Sharpe ratio (ours $1.133$ vs \textbf{Trans.} $1.028$). These results further demonstrate that the proposed method can be used in real-world investment to achieve higher risk-return rewards.
\end{itemize}

\subsection{Incremental Analysis}\label{sec:ana}
In this section, we want to answer the following research questions through incremental analysis:
\begin{itemize}
    \item \textbf{RQ1} Can TRA learn multiple trading patterns and select the correct one for prediction?
    \item \textbf{RQ2} Are both latent representation and temporal prediction errors useful for the router?
    \item \textbf{RQ3} How does optimal transport help the training of TRA?
    \item \textbf{RQ4} Is TRA sensitive to the number of $K$ predictors used?
\end{itemize}

\subsubsection{RQ1}
\begin{figure}[ht]
  \centering
   \includegraphics[width=1.0\columnwidth]{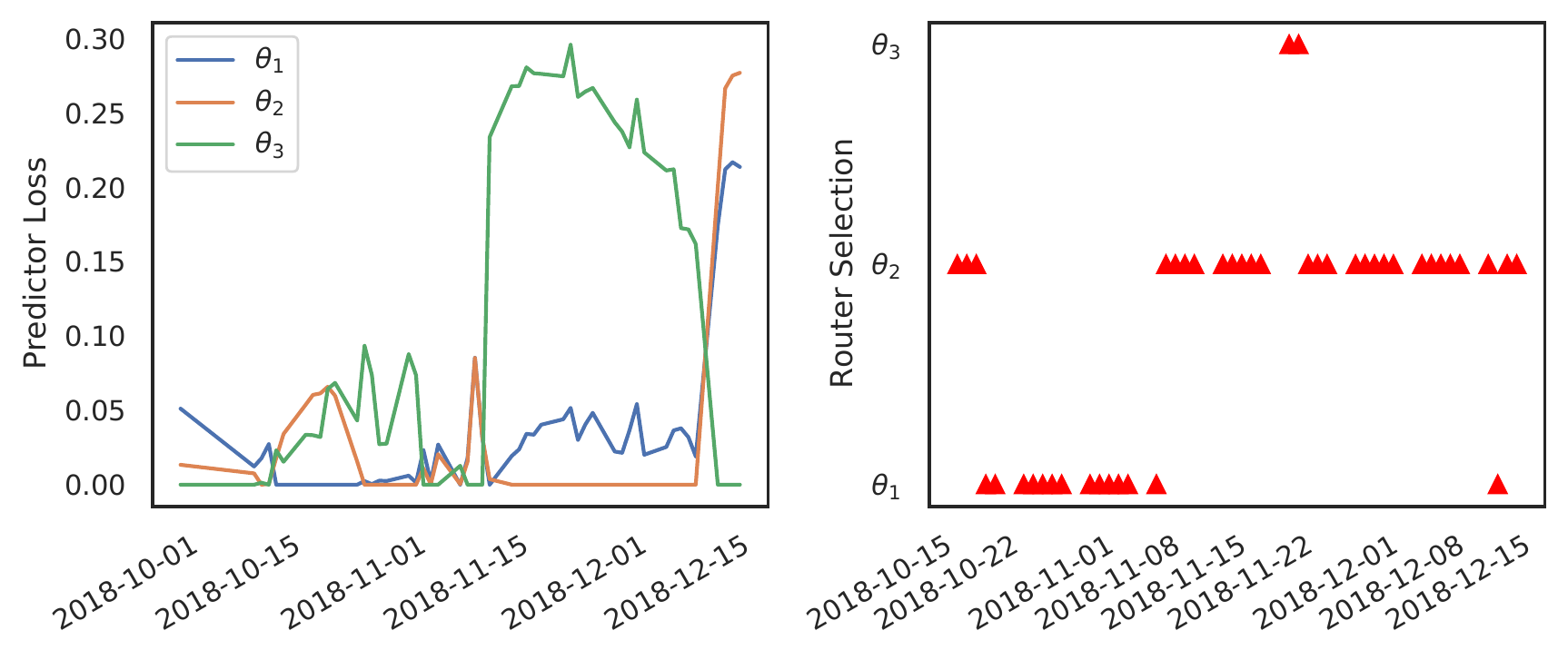}
  \caption{Loss of different predictors (left) and the router's selection (right). Loss is normalized by subtracting the minimum value of three.}
  \label{fig:pattern}
\end{figure}

We first analyse the learned multiple predictors with $K=3$ by showing their prediction loss of a randomly selected stock in Figure~\ref{fig:pattern}. It can be observed that the learned predictors have different performances at the same time, which demonstrates TRA does capture the diverse trading patterns. We also plot the selections given by the router in the right side and it can be observed that the router successfully picks the right model for prediction (the one with the lowest prediction loss). This evidence further justify the capability of TRA to select the correct model for inference.

\subsubsection{RQ2}
To understand what types of information can be used by the router to make good assignments of different samples, we compare the following information sources with ALSTM as the backbone model
\begin{itemize}
    \item \textbf{Random} simply uses Gaussian noise as the input information for the router.
    \item \textbf{Latent Representations, LR} are the extracted latent representation from the backbone feature extractor.
    \item \textbf{Temporal Prediction Errors, TPE} are prediction errors of different predictors defined in Equation~\ref{eq:tpe}.
    \item \textbf{LR+TPE} combines \textbf{LR} and \textbf{TPE} as the input for the router.
\end{itemize}

The model performances when using the above different types of information are summarized in Table~\ref{tab:info}. We can see that compared to \textbf{Random}, both \textbf{LR} and \textbf{TPE} consistently improve TRA's performance and the best performance can be achieved when both \textbf{LR} and \textbf{TPE} are used in TRA.
\begin{table}[ht]
    \centering
    \caption{TRA's performance when used with different information sources.} \label{tab:info}
    \begin{tabular}{c|cccc}
		\toprule
        Information & MSE ($\downarrow$) & MAE ($\downarrow$) & IC ($\uparrow$)& ICIR ($\uparrow$)\\
		\midrule
		Random &  0.159 (0.001) &  0.321 (0.002) &  0.048 &  0.362 \\
        LR     &  0.158 (0.001) &  0.320 (0.001) &  0.053 &  0.409 \\
        TPE    &  0.158 (0.001) &  0.321 (0.001) &  0.049 &  0.381 \\
        \textbf{LR+TPE} &  \textbf{0.157 (0.000)} &  \textbf{0.318 (0.000)} &  \textbf{0.059} &  \textbf{0.460} \\
        \bottomrule
    \end{tabular}
\end{table}

\subsubsection{RQ3}
\begin{figure}[ht]
  \centering
   \includegraphics[width=0.6\columnwidth]{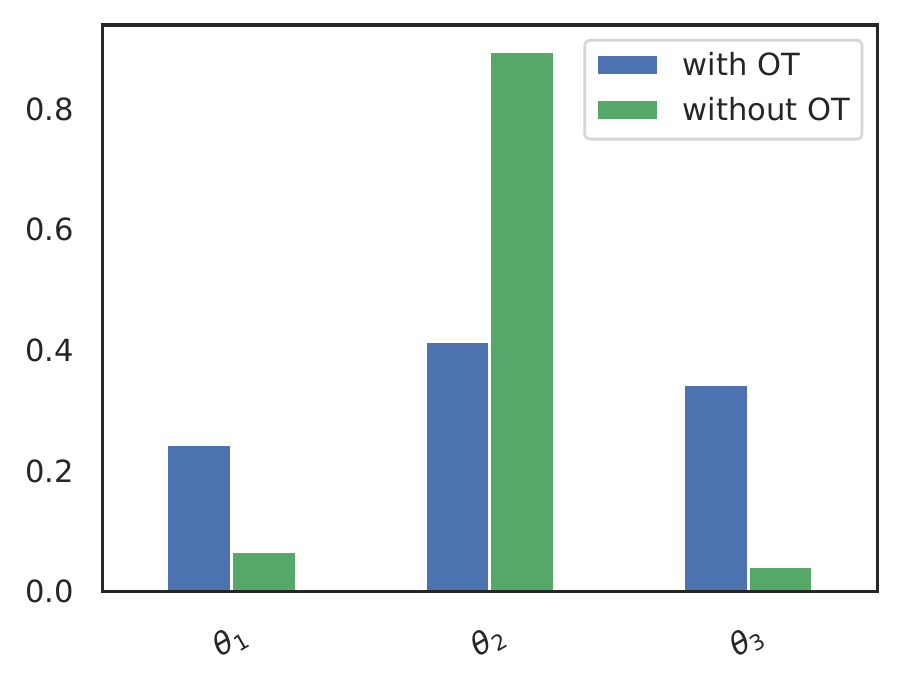}
  \caption{The relative shares of samples assigned to different predictors trained with or without optimal transport (OT).}
  \label{fig:sink}
\end{figure}

We now investigate the role of optimal transport (OT) used in our method. In our framework, we formalize the quest of optimal sample to predictor assignment task as an OT problem in Equation~\ref{eq:transport}, and use the solution from OT to guide the learning of the router through a regularization term in Equation~\ref{eq:loss}. We demonstrate the value of OT by comparing it with the naive objective without the regularization term. As illustrated in Figure~\ref{fig:sink}, after removing the OT regularization, the router tends to give some trivial assignment, where almost all samples are assigned to the same predictor and the remaining predictors are left unused. With OT, the assigned samples are more balanced thus all predictors can be well trained to capture the diverse trading patterns.

\subsubsection{RQ4}
\begin{figure}[ht]
  \centering
   \includegraphics[width=1.0\columnwidth]{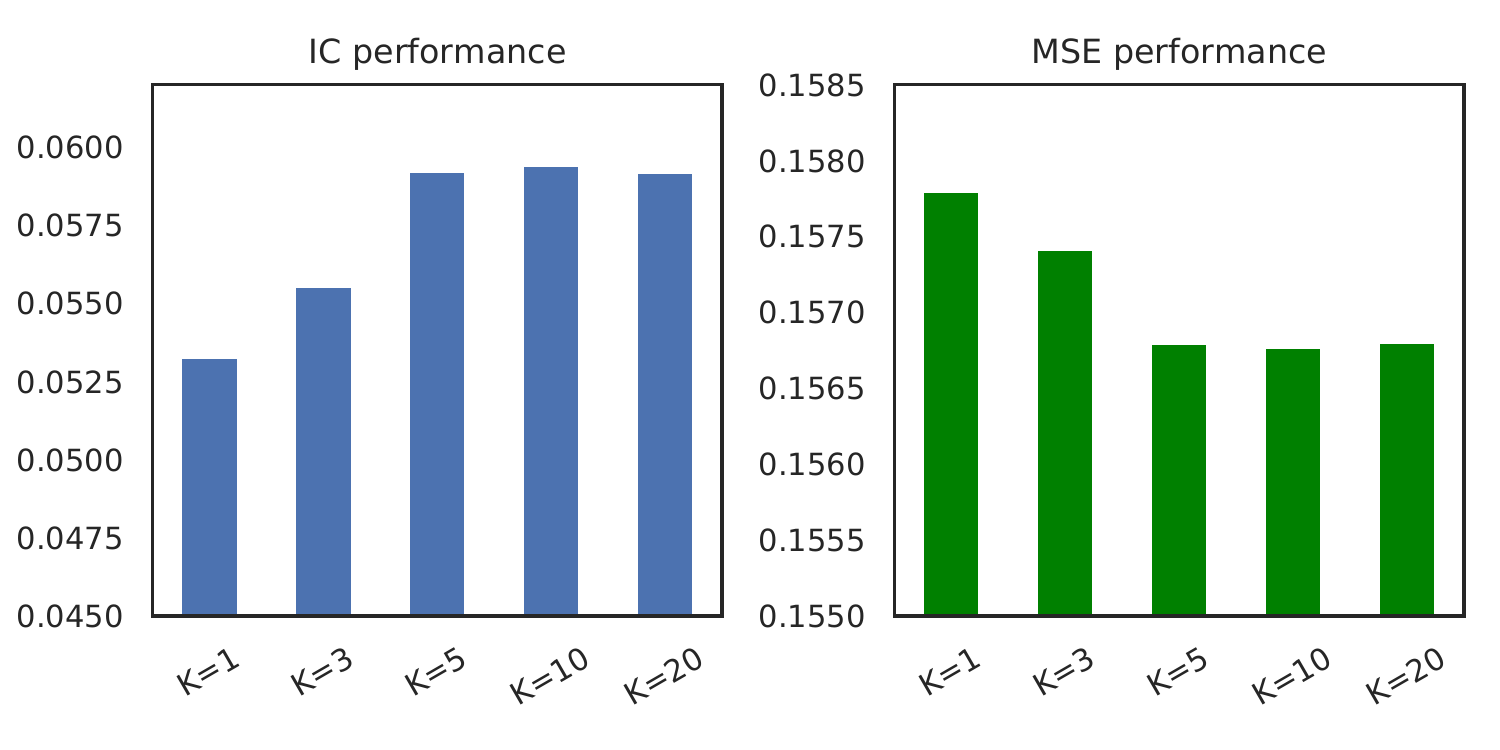}
  \caption{TRA's performance when using different number of predictors.}
  \label{fig:K_effect}
\end{figure}

Recall that TRA relies on a set of $K$ predictors to model multiple patterns where $K$ represents the number of latent patterns. As shown in Figure~\ref{fig:K_effect}, we compare TRA's performance measured by both IC and MSE when using different $K$s. It can be observed that compared to $K=1$ (i.e., only use the backbone model), introducing more predictors will consistently lead to better model performance. Specifically, adding four predictors ($K=5$) can bring huge performance gains, while adding more predictors can achieve slightly better performance. This suggests that a moderate selection of $K$ should give desirable performance.

\section{Conclusion}
In this paper, we show how the stock prediction performance can be further improved by learning multiple trading patterns in the stock market data. Specifically, We propose Temporal Routing Adaptor (TRA), which is a lightweight extension module that can be used to empower existing stock prediction networks with the ability of modeling multiple trading patterns. In order to encourage the discovery of the diverse trading patterns, we solve the sample to predictor assignment task as an Optimal Transport (OT) problem and use the solution from OT to guide the learning of the router. Experiments results demonstrate the effectiveness of the proposed method for learning multiple patterns and enhancing the stock prediction performance. Future work will take into account the new patterns emerging in the online environment.

\bibliographystyle{ACM-Reference-Format}
\bibliography{reference}

\appendix
\clearpage

\section{Portfolio Metrics} \label{sec:port_metrics}
In this paper we consider constructing a long-short portfolio to evaluate the predicted stock ranks from different models. Denote the selected stocks for the target portfolio on day $d$ as $\Lambda_d^+$ and $\Lambda_d^-$ for the long side and short side respectively, and denote the realized return for stock $s$ at day $d$ as $r_{s,d}$, where $r_{s,d}=0$ if $d$ is not a trade day (e.g., weekends). After carrying out trading simulations day by day, we will have a sequence of portfolio returns $\mathbf{R}=[R_1, R_2, ..., R_D]$ for $D$ days in the evaluation period, with each day's portfolio return $R_d$ calculated by
\begin{equation}
    R_d = \frac{1}{|\Lambda_d^+|} \sum_{s \in \Lambda_d^+} r_{s,d} - \frac{1}{|\Lambda_d^-|} \sum_{s \in \Lambda_d^-} r_{s,d}, \quad \forall d=1,2,...,D.
\end{equation}

Based on the portfolio returns sequence $\mathbf{R}$, there are quite a few metrics concerning different dimensions for how well this dynamic portfolio behaves. Among them, four metrics are used in this paper to evaluate the portfolio performance
\begin{itemize}
    \item \textbf{Annualized Return, AR.} Annualized Return measures the total profits generated by following the model predictions for investment on a yearly basis. It's calculated by scaling daily average portfolio return to the total number of calendar days in one year
    \begin{equation}
        \mathrm{AR} = mean(\mathbf{R}) \times 365 \times 100\%.
    \end{equation}

    \item \textbf{Annualized Volatility, AV.} Annualized Volatility measures the risk of the portfolio on a yearly basis. It's calculated by first calculating the standard deviation of daily portfolio returns, then scaling the deviation by the square root of the total number of calendar days in one year
    \begin{equation}
        \mathrm{AV} = std(\mathbf{R}) \times \sqrt{365} \times 100\%.
    \end{equation}

    \item \textbf{Sharpe Ratio, SR.} Sharpe Ratio is a measure of risk-adjusted return. It describes how much excess return we can get for the volatility risk we bear. SR can be calculated as the ratio of AR and AV
    \begin{equation}
        \mathrm{SR} = \frac{AR}{AV}.
    \end{equation}

    \item \textbf{Maximum Drawdown, MDD.} Maximum drawdown measures the worst performance of the portfolio in a given period. It is calculated as the maximum observed loss from a peak to a trough of a portfolio
    \begin{equation}
        \mathrm{MDD} = \max_{d\in[1,D]} \big\{ \max_{t \in [1, d]} \sum_{\le t} R_t - \sum_{\le d} R_d \big\}
    \end{equation}
\end{itemize}

\newpage
\section{Model Architecture and Hyper-parameters} \label{sec:arch}
Here are the detailed search space for model architecture and hyper-parameters:
\begin{itemize}
    \item \textbf{LightGBM}: \begin{itemize}
        \item \emph{number of estimators}: $500$
        \item \emph{number of leaves}: $[32, 64, 128, 256]$
        \item \emph{learning rate}: $[0.1, 0.05, 0.01]$
    \end{itemize}
    \item \textbf{MLP/SFM/ALSTM}: \begin{itemize}
        \item \emph{number of layers}: $[1, 2, 4]$
        \item \emph{hidden size}: $[32, 64, 128]$
        \item \emph{dropout}: $[0.0, 0.1, 0.2, 0.5]$
        \item \emph{learning rate}: $[0.001, 0.0005, 0.0002, 0.0001]$
        \item \emph{batch size}: $[512, 1024, 4096]$
    \end{itemize}
    \item \textbf{Transformer}: \begin{itemize}
        \item \emph{number of attention heads}: $[2, 4, 8]$
        \item other parameters are the same as MLP etc.
    \end{itemize}
\end{itemize}

\end{document}